\documentclass[conference]{IEEEtran}
\usepackage{amsmath,amssymb,amsfonts}
\usepackage{algorithmic,microtype}
\usepackage{graphicx}
\usepackage{dsfont}
\usepackage{textcomp}
\usepackage{xcolor}
\def\BibTeX{{\rm B\kern-.05em{\sc i\kern-.025em b}\kern-.08em
    T\kern-.1667em\lower.7ex\hbox{E}\kern-.125emX}}
\usepackage[0]{editing}

\DeclareMathAlphabet\calbf{OMS}{cmsy}{b}{n}

\begin{document}

\title{\huge Study of Energy-Efficient Distributed RLS-based Learning with Coarsely Quantized Signals \vspace{-0.5em}}

\author{\IEEEauthorblockN{Alireza Danaee$^{\star}$, \qquad Rodrigo C. de Lamare$^{\star,\dagger}$, \qquad and \qquad Vitor H. Nascimento$^{\ddagger}$}
\IEEEauthorblockA{$^{\star}$ Centre for Telecommunications Studies, Pontifical Catholic University of Rio de Janeiro, Brazil \\
$^{\dagger}$ Department of Electronic Engineering, University of York, United Kingdom \\
$^{\ddagger}$ Department of Electronic Systems Engineering, University of São Paulo, Brazil \\
danaee.alireza@gmail.com, delamare@cetuc.puc-rio.br, vitor@lps.usp.br\\{\footnotesize This work was supported in part by the ELIOT project, FAPESP 2018/12579-7 and  ANR-18-CE40-0030}
}}

\maketitle

\begin{abstract}
In this work, we present an energy-efficient distributed learning framework using coarsely quantized signals for Internet of Things (IoT) networks. In particular, we develop a distributed quantization-aware recursive least squares (DQA-RLS) algorithm that can learn parameters in an energy-efficient fashion using signals quantized with few bits while requiring a low computational cost. Numerical results assess the DQA-RLS algorithm against existing techniques for a distributed parameter estimation task where IoT devices operate in a peer-to-peer mode.
\end{abstract}


\section{Introduction}

Distributed signal processing algorithms are of great relevance for statistical inference in wireless networks and applications such as wireless sensor networks (WSNs) \cite{predd2006distributed} and the Internet of Things (IoT) \cite{rana2018iot}. In fact, distributed signal processing techniques deal with the extraction of information from data collected at nodes that are distributed over a geographical area. In this context, for each node a set of neighbor nodes collects and processes their local information, and transmits their estimates to a specific node. Then, each specific node combines the collected information together with its local estimate to generate improved estimates.

Prior work on distributed signal processing techniques has studied
protocols for exchanging information
\cite{olfati2007,lopes2008diffusion,cattivelli2008diffusion},
adaptive learning algorithms \cite{jio,jidf,xu2016distributed}, the
exploitation of sparse measurements and low-rank strategies
\cite{dce,miller2015sparsity,damdc,dlrest}, robust approaches
\cite{rdrls,dmlms,lrcc} and topology adaptation
\cite{xu2015adaptive}. Even though there have been many studies that
have evaluated the need for data exchange and signaling among nodes
as well as their computational complexity, prior work on
energy-efficient techniques is rather limited and there is no
distributed learning algorithm devised to deal with coarsely
quantized signals.

In this context, energy-efficient signal processing techniques have
gained a great deal of interest in the last decade or so due to
their ability to save energy and promote sustainable development of
electronic systems and devices. Electronic devices often exhibit an
energy consumption that is strongly dependent on the
analog-to-digital converters (ADCs) and the number of bits used to
represent digital samples \cite{walden1999analog}. This is of
central importance to devices that are battery operated and wireless
networks that must keep the energy consumption to a low level for
sustainability reasons. In particular, prior work on energy
efficiency has reported many contributions in signal processing for
communications and electronic systems that operate with coarsely
quantized signals
\cite{jacobsson2017throughput,mezghani2007modified,bbprec,1bitidd,1bitcpm,dynover}.

In this work, we propose an energy-efficient distributed learning
framework using low-resolution ADCs and signals for IoT networks. In
particular, we devise a distributed quantization-aware recursive
least squares (DQA-RLS) algorithm that can learn parameters in an
energy-efficient way using signals quantized with few bits with a
low computational cost and outperform the distributed
quantization-aware least-mean square (DQA-LMS) algorithm
\cite{dqa-lms}. Simulations assess the proposed DQA-RLS algorithm
against existing techniques for a distributed parameter estimation
task where IoT devices operate in a peer-to-peer mode.

This paper is structured as follows: Section 2 introduces the signal
model and states the problem. Section 3 details the proposed DQA-RLS
algorithms, whereas section 4 shows and discusses the results of
simulations and section 5 draws the conclusions of this work.

\section{Signal Model and Problem Statement}

\begin{figure}[htbp]
    \centering
    \includegraphics[width=8.0cm]{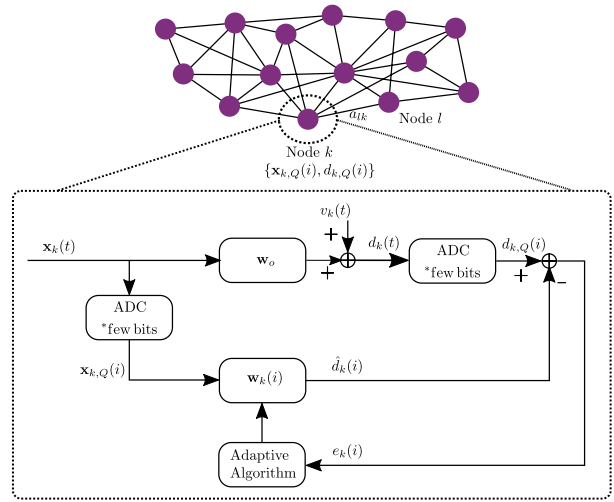}
    \caption{A distributed adaptive IoT network}
    \vspace{-0.8em}
    \label{dnet}
\end{figure}
We consider an IoT network consisting of $N$ nodes or agents which runs distributed signal processing techniques to perform the desired tasks, as depicted in Fig.~\ref{dnet}. The model adopted considers a desired signal $d_{k}(i)$, at each time $i$ described by
\begin{equation}
        d_k(i) = {\mathbf w}_o^H {\mathbf x}_{k}(i)+v_{k}(i), \quad k=1,2,\dots,N  \label{regres},
\end{equation}
where ${\mathbf w}_o \in \mathbb{C}^{M \times 1}$ is the parameter vector that the agents must estimate, ${\mathbf x}_k(i) \in \mathbb{C}^{M \times 1}$ is the regressor at node $k$, and $v_{k}(i)$ represents Gaussian noise with zero mean and variance $\sigma_{v,k}^2$ at each node $k$. We also consider the Adapt-then-Combine (ATC) diffusion rule as a more effective scheme than other previously reported schemes such as incremental and consensus \cite{olfati2007,lopes2008diffusion}.

As shown in Fig.~\ref{dnet}, because the measurement data at each node and the unknown system are analog and each agent processes the local data \{$d_k(i)$, ${\mathbf x}_k(i)$\} digitally, we need two ADCs in each agent. One concern is that as the number of agents increases, the energy consumption will grow too much when using high-resolution ADCs for each agent. This motivates us to quantize signals using few bits. Therefore, the problem we are interested in solving in this work is how to design energy-efficient distributed learning algorithms that can cost-effectively operate with coarsely quantized signals.

\section{Proposed DQA-RLS Algorithm}
\label{sec:proposed}

Let ${\mathbf x}_{k,Q}=Q_b({\mathbf x}_k)$ denote the $b$-bits quantized output of an ADC at node $k$, described by a set of $2^b+1$ thresholds ${\cal T}_b=\{\tau_0,\tau_1,...,\tau_{2^b}\}$, such that $-\infty=\tau_0<\tau_1<...<\tau_{2^b}=\infty$, and the set of $2^b$ labels ${\cal L}_b=\{l_0,l_1,...,l_{2^b-1}\}$ where $l_p \in (\tau_p,\tau_{p+1}]$, for $p \in [0,2^b-1]$ \cite{jacobsson2017throughput}. Let us assume that ${\mathbf x}_k\sim \mathcal{CN}({\mathbf 0},\,{\mathbf R}_{x_k})$ where ${\mathbf R}_{x_k} \in \mathbb{C}^{M \times M}$ is the covariance matrix of ${\mathbf x}_k$. We now use Bussgang's theorem \cite{bussgang1952crosscorrelation} to derive a model for the quantized vector ${\mathbf x}_{k,Q}$, which we later use to derive our DQA-RLS algorithm.  Employing Bussgang's theorem, ${\mathbf x}_{k,Q}$ can be decomposed as
\begin{align}
    {\mathbf x}_{k,Q}={\mathbf G}_{k,b}{\mathbf x}_k+{\mathbf q}_k \label{dcmps}
\end{align}
where the quantization distortion ${\mathbf q}_k$ is uncorrelated with ${\mathbf x}_k$, and ${\mathbf G}_{k,b} \in \mathbb{R}^{M \times M}$ is a diagonal matrix described by
\begin{equation}
    \begin{split}
        {\mathbf G}_{k,b} = {\rm diag}({\mathbf R}_{x_k})^{-\frac{1}{2}} \sum_{j=0}^{2^b-1} & \frac{l_j}{\sqrt{\pi}} \left[\exp(-\tau_j^2 {\rm diag} ({\mathbf R}_{x_k})^{-1}) \right.\\
        & \left.-\exp(-\tau_{j+1}^2 {\rm diag} ({\mathbf R}_{x_k})^{-1})\right].
    \end{split}  \label{Gk}\raisetag{25pt}
\end{equation}
Note that, as a simplifying approximation, we also apply this signal decomposition to the desired signal, $d_{k,Q}$, which is the output of the second ADC in the system, and for the particular case that ${\mathbf R}_{x_k}= \mathbb{E}[{\mathbf x}_k{\mathbf x}_k^H] = \sigma_{x,k}^2 {\mathbf I}_M$, the matrix ${\mathbf G}_{k,b}$ becomes $g_{k,b} {\mathbf I}_M$. However, to minimize the mean square error (MSE) between ${\mathbf x}_k$ and ${\mathbf x}_{k,Q}$, we need to characterize the probability density function (PDF) of ${\mathbf x}_k$ to find the optimal quantization labels. Because choosing these labels based on such PDF is ineffective in practice (since the PDFs are difficult to estimate), we assume the regressor ${\mathbf x}_{k}(i)$ is Gaussian, then adapt the approach in \cite{jacobsson2017throughput} and approximate the thresholds and labels as follows:
\begin{enumerate}
  \item We generate an auxiliary Gaussian random variable with unit variance and then use the Lloyd-Max algorithm \cite{lloyd1982least}, \cite{max1960quantizing} to find a set of thresholds ${\cal {\widetilde{T}}}_b=\{\tau_1,\dots,\tau_{2^b-1}\}$ and labels ${\cal {\widetilde{L}}}_b=\{\widetilde{l}_0,\dots,\widetilde{l}_{2^b-1}\}$ that minimize the MSE between the unquantized and the quantized signals.

  \item We wrap up the set of thresholds ${\cal T}_b$ by adding $\tau_0=-\infty$ and $\tau_{2^b}=\infty$ to the ${\cal {\widetilde{T}}}_b$.

  \item We rescale the labels such that the variance of the auxiliary random variable is 1. To do this, we multiply each label in the set ${\cal {\widetilde{L}}}_b$ by
  \begin{align}
      \alpha = \frac{1}{\sqrt{2 \sum\limits_{j=0}^{2^b-1} \tilde{l}_j^2(\Phi (\sqrt{2\tau_{j+1}^2})-\Phi (\sqrt{2\tau_{j}^2}))}}
  \end{align}
  to produce a set of suboptimal labels ${\cal L}_b = \alpha {\cal {\widetilde{L}}}_b$ , where $\Phi(.)$ is the cumulative distribution function (CDF) of a standard Gaussian random variable.
\end{enumerate}

We compute these thresholds and labels offline and use them to build the diagonal matrix ${\mathbf G}_{k,b}$ for the proposed DQA-RLS algorithm in what follows.

\subsection{Derivation of DQA-RLS}\label{AA}
\label{ssec:derv}

We consider ${\mathbf x}_k(t)$ and $d_k(t)$ as the analog input and output of the unknown system ${\mathbf w}_o$ at node $k$. Let ${\mathbf x}_k(i)$ and $d_k(i)$ denote the digital versions of ${\mathbf x}_k(t)$ and $d_k(t)$, and ${\mathbf x}_{k,Q}(i)$ and $d_{k,Q}(i)$ denote the coarsely quantized versions of ${\mathbf x}_k(i)$ and $d_k(i)$, respectively. We assume that the input signal at each node is  Gaussian with zero mean and covariance matrix ${\mathbf R}_{x_k}=E[{\mathbf x}_k {\mathbf x}_k^H]= \sigma_{x,k}^2 {\mathbf I}_M$ for $k=1,2,...,N$. We can now write $d_{k,Q}(i)$ as
\begin{equation}
    \begin{split}
        d_{k,Q}(i) &= Q(d_k(i)) = g_{k,b}(i)d_k(i)+q_k(i) \\
        \qquad &= g_{k,b}(i)( {\mathbf w}_o^H{\mathbf x}_k(i)+v_{k}(i))+q_k(i) \\
        \qquad &= g_{k,b}(i){\mathbf w}_o^H{\mathbf x}_k(i)+\hat{q}_k(i) \label{dcmpsD},
    \end{split}
\end{equation}
where $\hat{q}_k(i)=g_{k,b}(i)v_{k}(i)+q_k(i)$ and $g_{k,b}(i)$ is built from an estimate of ${\mathbf R}_{x_k}$ given by $\widehat{{\mathbf R}}_{x_k} ={\mathbf x}_k{\mathbf x}_k^H$ \cite{li2017channel}, as it depends on the choice of the input vector ${\mathbf x}_k$ due to \eqref{regres}. Because the adaptive algorithm receives a quantized signal, ${\mathbf x}_{k,Q}$, and the signal is assumed to be wide-sense stationary, at each time instant, we estimate $\sigma_{x,k}^2$ using the variance of the received input, $\sigma_{x_{k,Q}}^2$ and the distortion factor of the b-bit quantization, $\rho_{k,b}$, such that $\sigma_{x,k}^2 \approx \sigma_{x_{k,Q}}^2 + \rho_{k,b}$, where $\rho_{k,b} \approx \frac{\pi \sqrt{3}}{2} 2^{-2b}$  \cite{mezghani2007modified} for a Gaussian signal using non-uniform quantization in order to obtain the scalar $g_{k,b}(i)$.

Let us consider a network of $N$ nodes distributed over an area as in Fig.~\ref{dnet}. At time $i$, we collect the quantized desired signal and noise samples into vectors ${\mathbf d}_{i,Q}$ and ${\mathbf v}_i$, and the quantized input regressors into a matrix ${\mathbf X}_{i,Q}$ as follows
\begin{equation}
    \begin{split}
        {\mathbf X}_{i} &= {\rm col}\{{\mathbf x}_{1}^T(i),\dots,{\mathbf x}_{N}^T(i)\}   \qquad   (N \times M) \\
        {\mathbf X}_{i,Q} &= {\rm col}\{{\mathbf x}_{1,Q}^T(i),\dots,{\mathbf x}_{N,Q}^T(i)\}   \qquad   (N \times M) \\
        {\mathbf d}_{i,Q} &= {\rm col}\{d_{1,Q}(i),\dots,d_{N,Q}(i)\}  \qquad   (N \times 1)\\
        {\mathbf v}_i &= {\rm col}\{v_1(i),\dots,v_N(i)\}  \qquad   (N \times 1) \\
        {\mathbf g}_{i,b} &= {\rm col}\{g_{1,b}(i),\dots,g_{N,b}(i)\}  \qquad   (N \times 1).
    \end{split}
\end{equation}
If ${\mathbf v}_i^*$ denotes the complex conjugate transpose of ${\mathbf v}_i$, we can write down the covariance matrix of the noise vector as follows
\begin{equation}
    {\mathbf R}_v = \mathbb{E}[{\mathbf v}_i {\mathbf v}_i^*] = {\rm diag} \{\sigma^2_{v_1},\ldots ,\sigma^2_{v_N}\} \qquad (N \times N).
\end{equation}
Now we collect these data from time $0$ to time $i$ as follows
\begin{equation}
    \begin{split}
        {\calbf X}_{i} &= {\rm col} \{{\mathbf X}_{i},\dots,{\mathbf X}_{0}\}, \\
        {\calbf X}_{i,Q} &= {\rm col} \{{\mathbf X}_{i,Q},\dots,{\mathbf X}_{0,Q}\}, \\
        {\calbf D}_{i,Q} &= {\rm col} \{{\mathbf d}_{i,Q},\dots,{\mathbf d}_{0,Q}\}, \\
        {\calbf V}_i &= {\rm col} \{{\mathbf v}_i,\dots,{\mathbf v}_0\}, \\
        {\calbf G}_{i,b} &= {\rm col} \{{\mathbf g}_{i,b},\dots,{\mathbf g}_{0,b}\},
    \end{split}
\end{equation}
and write down $ {\calbf R}_{v,i} = \mathbb{E}[{\calbf V}_i{\calbf V}_i^*] $. In order to devise a learning algorithm based on \eqref{dcmpsD}, it is  convenient to first define $\hat{d}_k(i) = g_{k,b}(i){\mathbf w}_{k}^H(i-1){\mathbf x}_{k,Q}(i)$. Then, we estimate ${\mathbf w}_o$ by solving the weighted, regularized least squares problem given by
\begin{equation}
    \min_{\mathbf w} \| {\mathbf w}-\overline{\mathbf w}\| ^2_{{\mathbf \Pi}_i}+\| {\calbf D}_{i,Q}-{\calbf G}_{i,b} {\mathbf w}^H {\calbf X}_{i,Q} \| ^ 2_{{\calbf W}_i} \label{minw}
\end{equation}
The common solution ${\mathbf w}(i)$ is given by \cite{sayed2003fundamentals}
\begin{equation}
    {\mathbf w}_i = \overline{\mathbf w} + ({\mathbf \Pi}_i + {\calbf X}_{i,Q}^* {\calbf W}_i {\calbf X}_{i,Q} )^{-1}  {\calbf X}_{i,Q}^* {\calbf W}_i ({\calbf D}_{i,Q} - {\calbf G}_{i,b} \overline{\mathbf w}^H {\calbf X}_{i,Q}) \label{wgen}
\end{equation}
where ${\mathbf \Pi}_i > 0$ and ${\calbf W}_i >0$ are the regularization and the weighting Hermitian matrices.
An exponentially weighted version of \eqref{minw} can be derived choosing
\begin{align}
    {\calbf W}_i &= {\calbf R}_{v,i}^{-1} {\mathbf \Lambda}_i & &\text{and} & {\mathbf \Pi}_i &= \lambda^{i+1} {\mathbf \Pi},
\end{align}
where $0<\lambda<1$, ${\mathbf \Pi} >0$ and ${\mathbf \Lambda}_i \triangleq {\rm diag} \{ {\mathbf I}_N , \lambda {\mathbf I}_N, \dots,\lambda ^i {\mathbf I}_N $\}. Usually, ${\mathbf \Pi} = \delta ^{-1} {\mathbf I}_M$ where $\delta >0$ is large.
Often, in least squares estimation, when the noise variances $\sigma^2_{v_k}$ are unknown, the weighting matrix ${\calbf W}_i$ is simply replaced by ${\calbf W}_i = {\mathbf \Lambda}_i$.

Choosing $\overline{\mathbf w}=0$ the estimation problem \eqref{minw} will be
\begin{equation}
    \begin{split}
        {\mathbf w}(i) &= \arg\min_{{\mathbf w}} \left\{ \lambda ^{i+1} \|{\mathbf w}\|_{\mathbf \Pi}^2 +\right. \\
        &\left.\sum_ {j=0}^{i}{\lambda^{i-j}} \sum_{l=1}^{N}{\frac{|d_{l,Q}(j)-g_{k,b}(i){\mathbf w}_{k}^H(i-1){\mathbf x}_{k,Q}(i)|}{\sigma^2_{v_l}}}\right\}. \raisetag{45pt}\label{minwgl}
    \end{split}
\end{equation}
 We then reformulate the global least squares problem in (\ref{minwgl}) to a local least squares problem as follows:
\begin{equation}
    \begin{split}
        {\mathbf h}_k(i) &= \arg\min_{{\mathbf w}} \left\{ \lambda ^{i+1} \|{\mathbf w}\|_{\mathbf \Pi}^2 +\right.
        \\ &\sum_ {j=0}^{i}{\lambda^{i-j}} \left.\sum_{l=1}^{N}{\frac{c_{lk}|d_{l,Q}(j)-g_{k,b}(i){\mathbf w}_{k}^H(i-1){\mathbf x}_{k,Q}(i)|}{\sigma^2_{v_l}}}\right\}, \label{minwn}\raisetag{45pt}
    \end{split}
\end{equation}
for weighting coefficients $c_{lk}$ such that
$$c_{lk}=0 \ {\rm if} \ l \notin \mathcal {N}_k \ {\rm and} \ c_{lk}>0 \ {\rm if} \ l \in \mathcal {N}_k.$$

The coefficients $c_{lk}$ can be incorporated into the weighting matrix of \eqref{minw} by replacing ${\calbf W}_i$ with
\begin{equation}
    {\calbf W}_{k,i} = {\calbf R}_{v,i}^{-1} {\mathbf\Lambda_i} {\rm diag} \{ {\mathbf C}_k, \mathbf{C}_k,\dots,{\mathbf C}_k\}
\end{equation}
where ${\mathbf C}_k = {\rm diag}\{{\mathbf C}_{e_k}\} $ of size $N \times N$ , and ${\mathbf e}_k$ is the $N \times 1$ vector with a unity entry in position $k$ and zeros elsewhere.

When there is no regularization (${\mathbf \Pi} = {\mathbf 0}$), the solution ${\mathbf h}_k(i)$ for \eqref{wgen} is described by
\begin{equation}
    {\mathbf h}_k(i) = ({\calbf X}_{i,Q}^* {\calbf W}_{k,i} {\calbf X}_{i,Q} )^{-1}  {\calbf X}_{i,Q}^* {\calbf W}_{k,i} {\calbf D}_{i,Q} \label{hnoreg}
\end{equation}
where
\begin{equation}
    {\calbf D}_{i,Q} = Q \left ({\mathbf w}_o^H {\calbf X}_{i} + {\calbf V}_i \right ) = {\calbf G}_{i,b}{\mathbf w}_o^H {\calbf X}_{i} + \hat{\calbf V}_i \label{dcolq}.
\end{equation}
We replace ${\calbf W}_{i}$ in \eqref{wgen} with ${\calbf W}_{k,i}$, ${\mathbf \Pi}_i= \lambda ^{i+1} {\mathbf \Pi}$, and $\overline{\mathbf w} = {\mathbf 0}$ to obtain the solution to \eqref{minwn}. It can be formulated as
\begin{equation}
    {\mathbf h}_k(i) = {\mathbf P}_k(i) {\calbf X}_{i,Q}^* {\calbf W}_{k,i} {\calbf D}_{i,Q} \label{hkpk}
\end{equation}
where
\begin{equation}
    {\mathbf P}_k(i) = (\lambda ^{i+1} {\mathbf \Pi} + {\calbf X}_{i,Q}^* {\calbf W}_{k,i} {\calbf X}_{i,Q} )^{-1} \label{pkshrt} .
\end{equation}
To form the recursion, we compute ${\mathbf P}_k(i)$ from ${\mathbf P}_k(i-1)$ considering
\begin{equation}
    \begin{split}
            {\mathbf P}_k^{-1}(i) &= \lambda [\lambda^i {\mathbf \Pi} + {\calbf X}_{Q}(i-1) ^* {\calbf W}_{k}(i-1) {\calbf X}_{Q}(i-1) ] + \\
            & \qquad {\mathbf X}_{i,Q}^* {\mathbf R}_{v}^{-1} {\mathbf C}_{k} {\mathbf X}_{i,Q} \\
            & = \lambda {\mathbf P}_k^{-1}(i-1) + {\mathbf X}_{i,Q}^* {\mathbf R}_{v}^{-1} {\mathbf C}_{k} {\mathbf X}_{i,Q} \\
            & =  \lambda {\mathbf P}_k^{-1}(i-1) + \sum_{l=1}^N \frac{c_{lk}}{\sigma_{v_l}^2} {\mathbf x}_{l}(i) {\mathbf x}_{l}^*(i) \label{pkminus} .
    \end{split}
\end{equation}
To implement the recursion with reduced complexity, we use a series of rank-one updates as follows
\begin{equation}
    \begin{split}
        & {\mathbf P}_k^{0}(i) \gets \lambda^{-1} {\mathbf P}_k(i-1) \\
        & {\rm For}\  l = 1 \ {\rm to} \ N, {\rm repeat} \\
        & \qquad {\mathbf P}_k^{l}(i) \gets \left [ \left ({\mathbf P}_k^{l-1}(i) \right)^{-1} +  \frac{c_{lk}}{\sigma_{v_l}^2} {\mathbf x}_{l}(i) {\mathbf x}_{l}^*(i) \right ] ^{-1} \\
        & {\rm end} \\
        & {\mathbf P}_k(i) \gets {\mathbf P}_k^{N}(i) \label{pkrecur1}.
    \end{split}
\end{equation}
The matrices ${\mathbf P}_k^{l}(i)$ denote the intermediate results after every rank-one update and are needed for only neighbor nodes where $c_{lk} \neq 0$. Using the matrix inversion lemma, we can write \eqref{pkrecur1} as follows
\begin{equation}
    \begin{split}
        & {\mathbf P}_k(i) \gets \lambda^{-1} {\mathbf P}_k(i-1) \\
        & {\rm For\ every}\ l \in \mathcal {N}_k \ {\rm repeat} \\
        & \qquad {\mathbf P}_k(i) \gets {\mathbf P}_k(i) - \frac{c_{lk} {\mathbf P}_k(i) {\mathbf x}_{l}(i){\mathbf x}_{l}^*(i){\mathbf P}_k(i)}{\sigma_{v_l}^2 + c_{lk}{\mathbf x}_{l}^*(i) {\mathbf P}_k(i){\mathbf x}_{l}(i)} \\
        & {\rm end} \label{pkrecur}.
    \end{split}
\end{equation}
To build a recursion for the update of \eqref{hkpk}, we collect all measurements and regressors from all nodes up to time $i-1$, from nodes $1$ to $l$ and denote the intermediate matrices as
\begin{equation}
    \begin{split}
        & {\calbf D}_{i,Q}^l = \begin{bmatrix} d_{l,Q}(i) \\ \vdots \\ d_{1,Q}(i) \\ \overline{{\calbf D}_{i-1,Q}} \end{bmatrix} \ {\rm ,} \ {\calbf X}_{i,Q}^l = \begin{bmatrix} {\mathbf x}_{l,Q}(i) \\ \vdots \\ {\mathbf x}_{1,Q}(i) \\ \overline{{\calbf X}_{i-1,Q}} \end{bmatrix} \\
         & {\calbf W}_{k,i}^l = \begin{bmatrix} \frac{c_{lk}}{\sigma_{v_l}^2} & & &\\ & \ddots & & \\ & & \frac{c_{1k}}{\sigma_{v_1}^2} & \\ & & & \lambda {\calbf W}_{k,i-1} \end{bmatrix} \label{xdint}.
    \end{split}
\end{equation}
This allows us to write the intermediate estimates ${\mathbf h}_k^l(i)$ as
\begin{equation}
    {\mathbf h}_k^l(i) = {\mathbf P}_k^l(i) ({\calbf X}_{i,Q}^l)^* {\calbf W}_{k,i}^l {\calbf D}_{i,Q}^l, \label{hkpkint}
\end{equation}
where the $\mathbf{P}_k^l(i)$ is defined in \eqref{pkrecur1}, and can also be written as
\begin{equation}
    {\mathbf P}_k^l(i) = [\lambda^{i+1} {\mathbf \Pi} + ({\calbf X}_{i,Q}^l)^* {\calbf W}_{k,i}^l {\calbf D}_{i,Q}^l]^{-1} \label{pkint}.
\end{equation}
It holds that ${\mathbf h}_k(i-1) = {\mathbf h}_k^N(i-1) \triangleq {\mathbf h}_k^0(i)$.  Let $j$ denote the smallest index such that $c_{jk} \neq 0$. We can calculate ${\mathbf h}_k^j(i)$ from ${\mathbf h}_k^0(i)$ as follows
\begin{equation}
    \begin{split}
        {\mathbf h}_k^j(i) &= {\mathbf P}_k^{j}(i) \left [ \lambda ({\calbf X}_{i,Q}^0)^* {\calbf W}_{k,i}^0 {\calbf D}_{i,Q}^0 + \frac{c_{jk}}{\sigma_{v_j}^2} {\mathbf x}_{j,Q}(i) d_{j,Q}(i) \right ] \\
        &= \underbrace { \lambda {\mathbf P}_k^0(i) ({\calbf X}_{i,Q}^0)^* {\calbf W}_{k,i}^0 {\calbf D}_{i,Q}^0}_{{\mathbf h}_k(i-1)} + \frac{c_{jk}}{\sigma_{v_j}^2} {\mathbf P}_k^0(i) {\mathbf x}_{j,Q}(i) \\
        &  \left ( 1- \frac{c_{jk} {\mathbf x}_{j}^*(i) {\mathbf P}_k^0(i) {\mathbf x}_{j}(i)}{\sigma_{v_j}^2 + c_{jk}{\mathbf x}_{j}^*(i) {\mathbf P}_k^0(i){\mathbf x}_{j}(i)} \right ) d_{j,Q}(i) \\
        &- \frac{c_{jk} {\mathbf x}_{j}^*(i) {\mathbf P}_k^0(i) {\mathbf x}_{j}(i)}{\sigma_{v_j}^2 + c_{jk}{\mathbf x}_{j}^*(i) {\mathbf P}_k^0(i){\mathbf x}_{j}(i)} \underbrace { \lambda {\mathbf P}_k^0(i) ({\calbf X}_{i,Q}^0)^* {\calbf W}_{k,i}^0 {\calbf D}_{i,Q}^0}_{{\mathbf h}_k(i-1)}  \\
        &= {\mathbf h}_k(i-1) + \frac{c_{jk} {\mathbf P}_k^0(i) {\mathbf x}_{j}(i)}{\sigma_{v_j}^2 + c_{jk}{\mathbf x}_{j}^*(i) {\mathbf P}_k^0(i){\mathbf x}_{j}(i)} \left( d_{j,Q}(i) \right. \\
        &\left. - g_{k,b}(i) {\mathbf h}_k^H(i-1) {\mathbf x}_{j}(i) \right).
    \end{split} \raisetag{12pt}\label{hintrec}
\end{equation}
We derive recursions for ${\mathbf h}_k^l(i)$ from ${\mathbf h}_k^{l-1}(i)$ for $l=j+1, \dots ,N$ in a similar way as
\begin{equation}
    \begin{split}
        {\mathbf h}_k^l(i) &= {\mathbf P}_k^l(i) \left [ ({\calbf X}_{i,Q}^{l-1})^* {\calbf W}_{k,i}^{l-1} {\calbf D}_{i,Q}^{l-1} + \frac{c_{lk}}{\sigma_{v_l}^2} {\mathbf x}_{l,Q}(i) d_{l,Q}(i) \right ] \\
        &= {\mathbf h}_k^{l-1}(i-1) + \frac{c_{lk} {\mathbf P}_k^{l-1}(i) {\mathbf x}_{l}(i)}{\sigma_{v_l}^2 + c_{lk}{\mathbf x}_{l}^*(i) {\mathbf P}_k^{l-1}(i){\mathbf x}_{l}(i)} \left[ d_{l,Q}(i) \right. \\
        &\left.- g_{k,b}(i)({\mathbf h}_k^{l-1})^H(i-1) {\mathbf x}_{l}(i) \right ].
    \end{split}\raisetag{12pt}\label{hlintrec}
\end{equation}
To simplify the recursion, we drop the super-indexes $j$ as we only need to consider values of $l$ which $c_{lk} \neq 0$. Then regarding all time instants up to time $i$, we combine \eqref{pkrecur}, \eqref{hintrec} and \eqref{hlintrec} to build the recursion to solve \eqref{minwn} as in Table \ref{PCode}.

We improve the estimation by exchanging the estimates between neighbor nodes in the diffusion fashion using a weighted average of the estimates of the nodes as in the last step in table \ref{PCode}.
The combination coefficients of neighbor nodes on node $k$, $a_{lk}$, are chosen such that
$$a_{lk}=0 \ {\rm if} \ l \notin \mathcal {N}_k, a_{lk}>0 \ {\rm if} \ l \in \mathcal {N}_k,\ {\rm and} \sum_{l\in{\mathcal{N}_k}} {a_{lk}} = 1.$$

\begin{table}[htb!]
\begin{small}
\caption{Pseudo code of DQA-RLS algorithm}
\begin{center}
\begin{tabular}{l}
\hline
Initializations: ${\mathbf w}_{k}(-1) = 0$ and ${\mathbf P}_k(-1) = {\mathbf {\mathbf \Pi}}^{-1}$ for each node $k$ \\
\hline
At each time instant $i$ and node $k$ repeat \\
\qquad $g_{k,b}(i) = \frac{1}{{\sqrt {\sigma_{x_k}^2}}} \sum\limits_{j=0}^{2^b-1} \frac{l_j}{\sqrt{\pi}} ( e^{- \frac{\tau_j^2}{\sigma_{x_k}^2}}-e^{- \frac{\tau_{j+1}^2}{\sigma_{x_k}^2}} )$ \\
\qquad ${\mathbf h}_k(i) = {\mathbf w}_k(i-1)$ \\
\qquad ${\mathbf P}_k(i) = \lambda^{-1} {\mathbf P}_k(i-1)$\\
\qquad for all $l \in {\mathcal N}_k$ \\
\qquad \qquad ${\mathbf h}_k(i) \gets {\mathbf h}_k(i) + \frac{c_{lk} {\mathbf P}_k(i) {\mathbf x}_{l}(i)[d_{l,Q}(i)-g_{k,b}(i){\mathbf h}_k^H(i) {\mathbf x}_{l}(i)] }{\sigma_{v_l}^2 + c_{lk}{\mathbf x}_{l}^*(i) {\mathbf P}_k(i){\mathbf x}_{l}(i)}$ \\
\qquad \qquad ${\mathbf P}_k(i) \gets {\mathbf P}_k(i) - \frac{c_{lk} {\mathbf P}_k(i) {\mathbf x}_{l}(i){\mathbf x}_{l}^*(i){\mathbf P}_k(i)}{\sigma_{v_l}^2 + c_{lk}{\mathbf x}_{l}^*(i) {\mathbf P}_k(i){\mathbf x}_{l}(i)}$ \\
\qquad end \\
\qquad for every node $k$ repeat \\
\qquad \qquad ${\mathbf w}_k(i)=\sum\limits_{l\in {\mathcal N}_k} a_{lk}{\mathbf h}_l(i)$ \\
\hline
\end{tabular}
\label{PCode}
\end{center}
\end{small}
\end{table}

\subsection{Complexity and Energy Consumption}\label{AB}
\label{ssec:compx}
\begin{figure}[htbp]
    \centering
    \includegraphics[width=8cm]{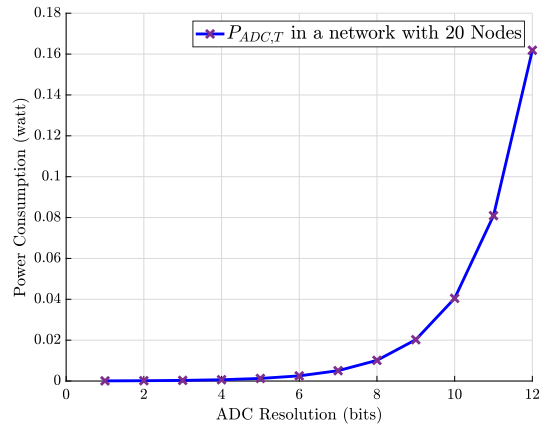}
    \caption{Power consumption of the ADCs in an adaptive IoT network.}
    \vspace{-1em}
    \label{Padc1}
\end{figure}

At each time instant, to compute ${g}_{k,b}(i)$ from \eqref{Gk}, $2^{b+1}+1$ multiplications, $2^b -1$ additions, $2^b +1$ divisions, and $2^b$ exponentiations are needed. Note that we compute $g_{k,b}(i)$ online, since this is more appropriate to deal with non-stationary input data. However, one can compute ${\mathbf G}_{k,b}$ offline having the covariance matrix of the input signal as in \eqref{Gk}. To compute ${\mathbf h}_k(i)$ in table \ref{PCode}, one more multiplication than in DRLS \cite{cattivelli2008diffusion} is needed. Therefore, DQA-RLS performs a few more operations ($\approx O(2^b)$) than DRLS.
However, the extra complexity in the DQA-RLS algorithm allows the system to work in a more energy-efficient way. In order to assess the power savings by low resolution quantization, let us consider a network with $N$ nodes in which each node uses two ADCs. The power consumption of each ADC is $P_{ADC}(b) = cB 2^b$ \cite{orhan2015low}, where $B$ is the bandwidth (related to the sampling rate), $b$ is the number of quantization bits of the ADC, and $c$ is the power consumption per conversion step. Therefore, the total power consumption of the ADCs in the network is
\begin{equation}
    P_{ADC,T}(b) = 2 N cB 2^b \qquad  ({\rm watts}). \label{Padcn}
\end{equation}
Fig.~\ref{Padc1} shows an example of the total power consumption of ADCs in a narrowband IoT (NB-IoT) network running diffusion adaptation consisting of 20 nodes with bandwidth $B=200 \, {\rm kHz}$ \cite{ratasuk2016nb} and considering the energy consumption per conversion step of each ADC, $c=494~ {\rm fJ}$, as in \cite{chung20097}.

\section{Simulation Results}
\label{sec:sims}
\begin{figure}[htbp]
\begin{minipage}[b]{.45\linewidth}
\centering \centerline{\includegraphics[width=4.75cm]{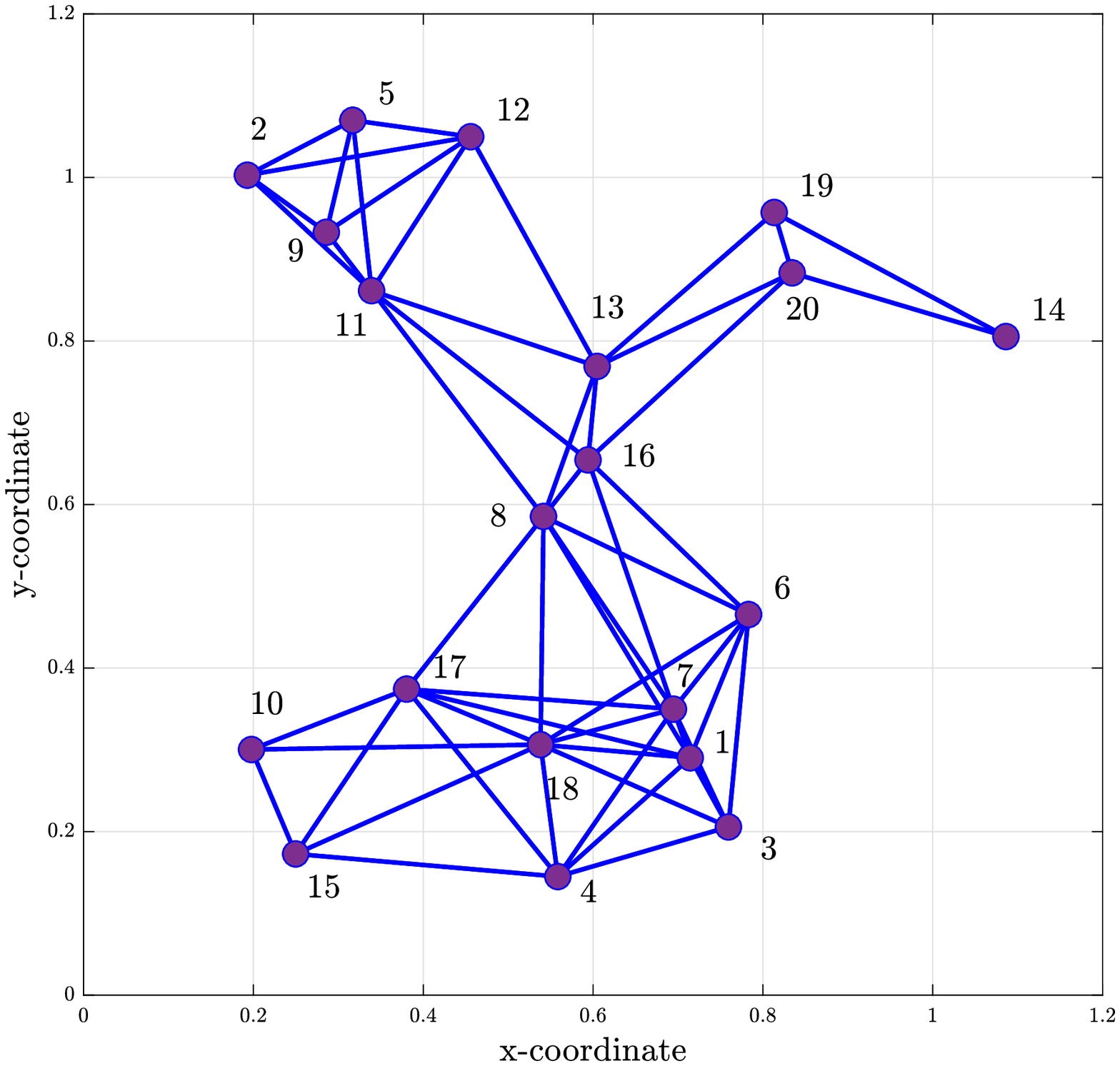}}
\centerline{(a) Distributed network structure}\medskip
\end{minipage}
\hfill
\begin{minipage}[b]{0.45 \linewidth}
\centering \centerline{
\includegraphics[width=4.75cm]{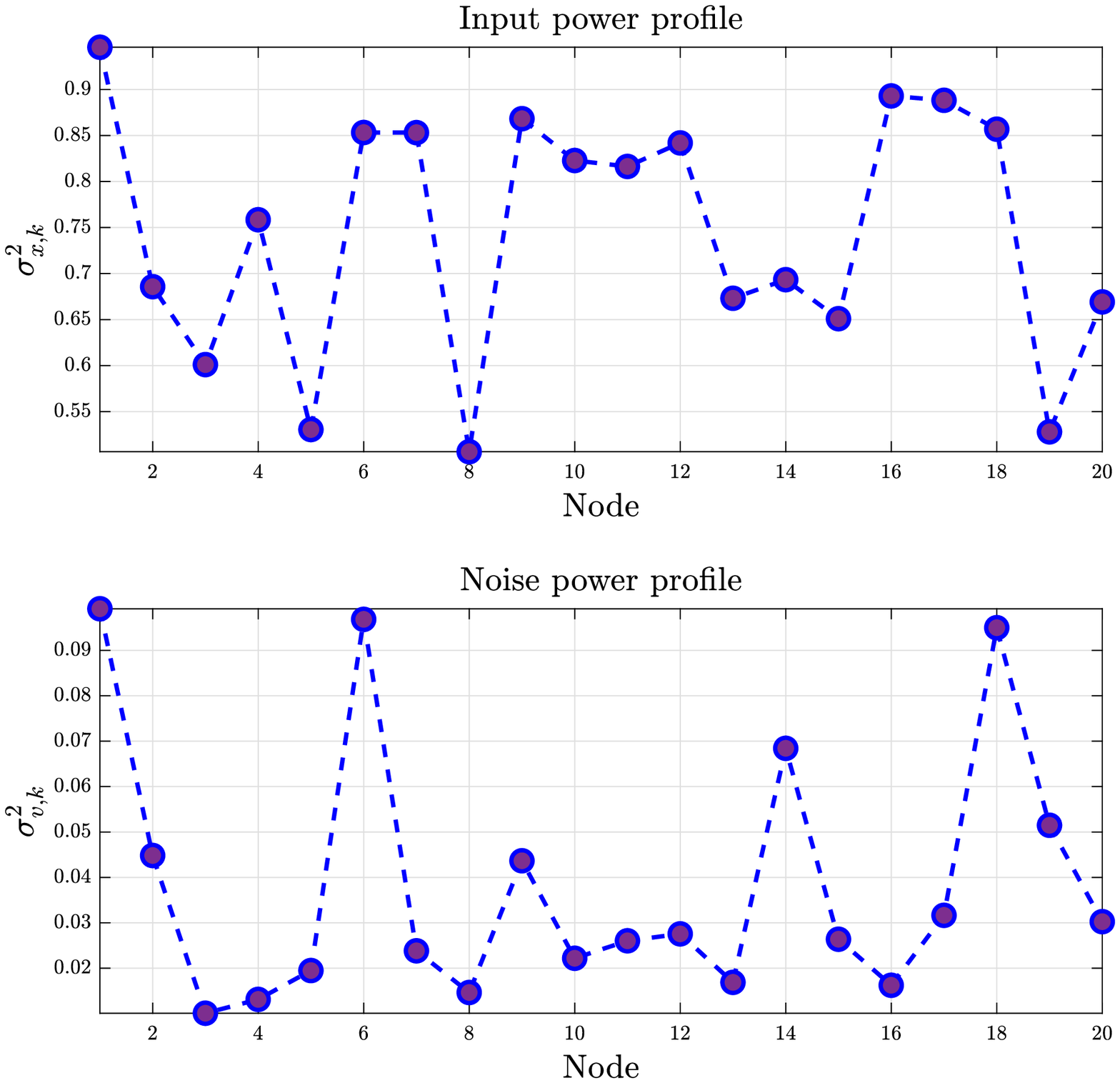}} \centerline{(b)
Input and noise variances}\medskip
\end{minipage}
\vspace{-1.65em}
\caption{A wireless network with $N=20$ nodes.}
\label{fig:netw}
\end{figure}

In this section, we assess the estimation performance of the DQA-RLS algorithm for a system identification setup in a network with $N=20$ nodes. The impulse response of the unknown system has $M=8$ taps, is generated randomly and normalized to one. The input signals ${\mathbf x}_k(i)$ at each node are generated by a white Gaussian noise process with variance $\sigma_{x,k}^2$ and quantized using Lloyd-Max quantization scheme to generate ${\mathbf x}_{k,Q}(i)$. The noise samples of each node are drawn from a zero mean white Gaussian process with variance $\sigma_{v,k}^2$. Fig. ~\ref{fig:netw} plots the network structure and the node profiles.

\begin{figure}[htbp]
\centerline{\includegraphics[width=8.25cm]{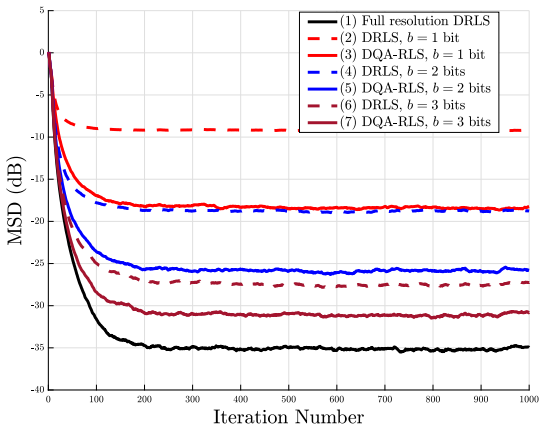}}
\vspace{-1.5em} \caption{The MSD curves for the DRLS and DQA-RLS
algorithms.} \vspace{-1.25em} \label{msd}
\end{figure}

The simulated mean-square deviation (MSD) learning curves are obtained by ensemble averaging over 100 independent trials. The combining coefficients $a_{lk}$ are computed by the Metropolis rule and $\lambda=0.98$. Fig.~\ref{msd} shows the MSD results obtained from simulations for DRLS and DQA-RLS using different numbers of bits. Curve 1 shows the standard DRLS performance assuming full resolution ADCs to perform system identification. Curves 2, 4 and 6 show the MSD evolution of the standard DRLS with signals coarsely quantized with b=1, 2 and 3 bits, respectively. Curves 3, 5 and 7 show the MSD performance of the proposed DQA-RLS algorithm that improves the MSD performance for coarsely quantized signals. The performance of the proposed DQA-RLS algorithm is closer to the DRLS while its energy consumption is over $90\%$ less than that of the standard DRLS with full resolution (see Fig.~\ref{Padc1}).

\section{Conclusion}
\label{sec:conc}

In this paper, we have proposed an energy-efficient framework for
distributed learning and developed the DQA-RLS algorithm for
adaptive IoT networks. We have also investigated the DQA-RLS
algorithm using low resolution ADCs. The proposed DQA-RLS algorithm
has comparable computational complexity to the standard DRLS
algorithm while it greatly reduces the power consumption of the ADCs
in the network. Simulations have shown the good performance of
DQA-RLS as compared to the standard DRLS for coarsely quantized
signals.

\bibliographystyle{IEEEbib}
\bibliography{refs}

\end{document}